\documentclass[journal]{IEEEtran}
\usepackage[dvipsnames, svgnames,table, x11names]{xcolor}
\ifCLASSINFOpdf
\else
   \usepackage[dvips]{graphicx}
\fi
\usepackage{url}

\usepackage{color}
\usepackage{times}
\usepackage{epsfig}
\usepackage{graphicx}
\usepackage{amsmath}
\usepackage{amssymb}
\usepackage{multirow}
\usepackage{mathrsfs}
\usepackage{array}
\usepackage{ifpdf}
\usepackage{algorithmic}
\usepackage{algorithm}
\usepackage{booktabs}
\usepackage{float}

\usepackage{cite}
\usepackage{subcaption}
\usepackage{changes}
\usepackage{diagbox}
\usepackage{multirow}
\usepackage[colorlinks,
            linkcolor=red,
            anchorcolor=blue,
            citecolor=green 
            ]{hyperref}
            
\begin{document}

\title{MirrorDiffusion: Stabilizing Diffusion Process in Zero-shot Image Translation by Prompts Redescription and Beyond}

\author{Yupei Lin, Xiaoyu Xian, Yukai Shi$^\dagger$, and Liang Lin, \IEEEmembership{Fellow, IEEE}
\thanks{Y. Lin, Y. Shi are with School of Information Engineering, Guangdong University of Technology, Guangzhou, 510006, China
       (email:  {\tt\small yupeilin2388@gmail.com; ykshi@gdut.edu.cn}).}%
       
\thanks{X. Xian is with CRRC Academy Co., Ltd.,
       (email:  {\tt\small xxy@crrc.tech}).}%
       
\thanks{L. Lin is with School of Computer Science, Sun Yat-sen University, Guangzhou, 510006, China
       (email:  {\tt\small linlng@mail.sysu.edu.cn}).}%
}


\twocolumn[{
\renewcommand\twocolumn[1][]{#1}
\maketitle
\begin{center}
    \captionsetup{type=figure}
    \includegraphics[width=.9\textwidth]{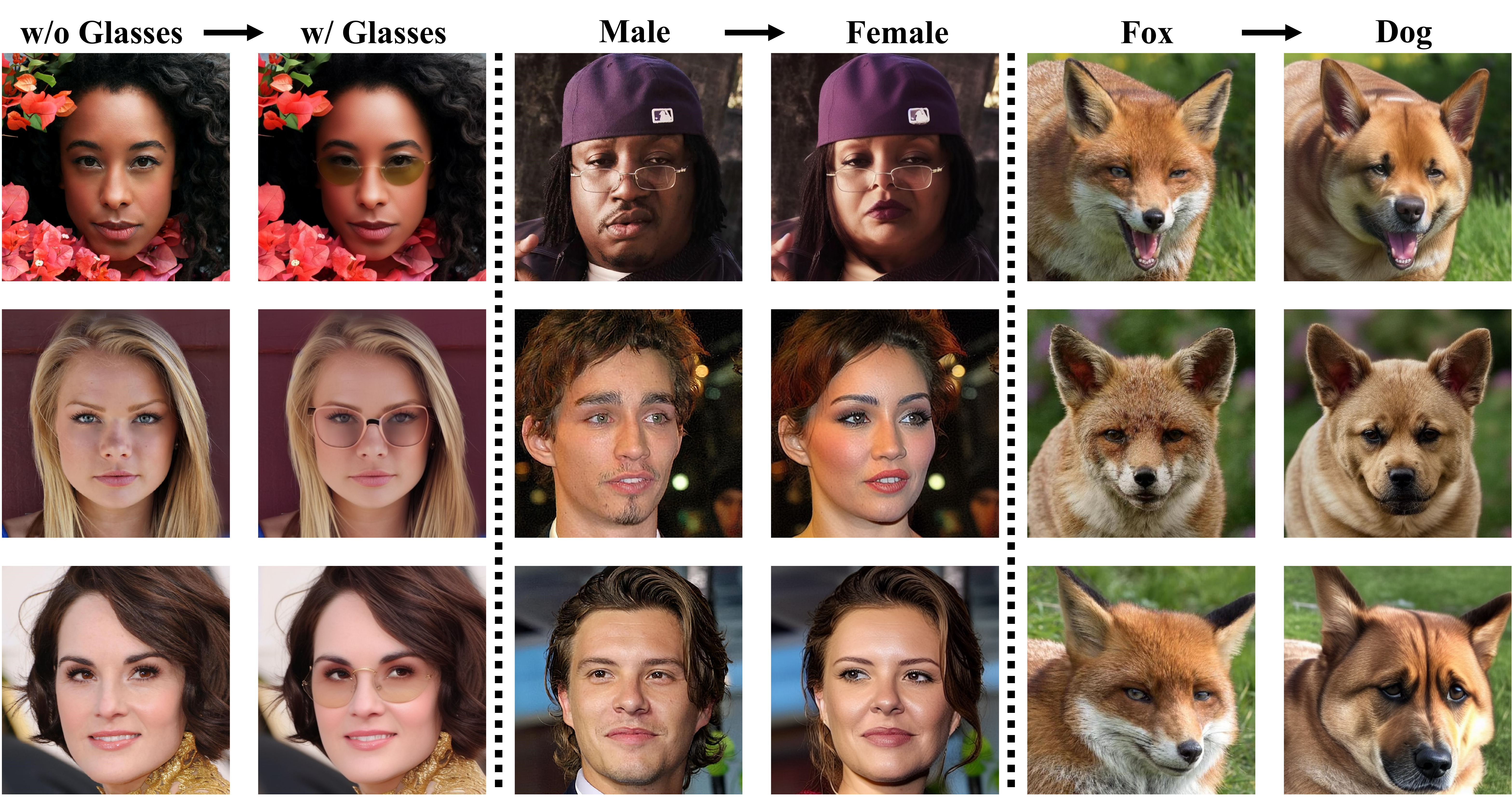}
    \captionof{figure}{Without any supervision, MirrorDiffusion realized three zero-shot image-to-image translations: w/o glasses $\rightarrow$ w/ glasses, Male $\rightarrow$ Female and Fox $\rightarrow$ Dog.}
    \label{fig:add_exp}
\end{center}
}]

\begin{abstract}

Recently, text-to-image diffusion models become a new paradigm in image processing fields, including content generation, image restoration and image-to-image translation. Given a target prompt, Denoising Diffusion Probabilistic Models (DDPM) are able to generate realistic yet eligible images. With this appealing property, the image translation task has the potential to be free from target image samples for supervision. By using a target text prompt for domain adaption, the diffusion model is able to implement zero-shot image-to-image translation advantageously. However, the sampling and inversion processes of DDPM are stochastic, and thus the inversion process often fail to reconstruct the input content. Specifically, the displacement effect will gradually accumulated during the diffusion and inversion processes, which led to the reconstructed results deviating from the source domain. To make reconstruction explicit, we propose a prompt redescription strategy to realize a mirror effect between the source and reconstructed image in the diffusion model (MirrorDiffusion). More specifically, a prompt redescription mechanism is investigated to align the text prompts with latent code at each time step of the Denoising Diffusion Implicit Models (DDIM) inversion to pursue a structure-preserving reconstruction. With the revised DDIM inversion, MirrorDiffusion is able to realize accurate zero-shot image translation by editing optimized text prompts and latent code. Extensive experiments demonstrate that MirrorDiffusion achieves superior performance over the state-of-the-art methods on zero-shot image translation benchmarks by clear margins and practical model stability. Our project is available at \href{https://mirrordiffusion.github.io/}{https://mirrordiffusion.github.io/}

\end{abstract}


\IEEEpeerreviewmaketitle

\section{Introduction}

\IEEEPARstart{R}{ecently}, text-to-image diffusion model~\cite{song2020denoising} becomes a new fashion in signal processing fields. With large-scale pre-training on text-to-image data pairs, Denoising Diffusion Probabilistic Models (DDPM) have a strong capacity to generate diverse image content~\cite{saharia2022palette,ruiz2023dreambooth,ramesh2022hierarchical}. Nevertheless, DDPM has achieved great success in image generation task, it still fail to achieve a desired performance on image-to-image translation, especially on zero-shot image-to-image translation. 

Image translation~\cite{wang2018thermal,chen2022unsupervised} aims to transform images from a source domain to a target domain, such as cat $\to$ dog. This task inherently requires the target domain images for model adaption, which used to be accomplished by Generative Adversarial Nets (GAN)~\cite{goodfellow2014generative,xiang2019single,chen2020learning,hong2019unsupervised,yin2022laplacian,zhou2022searching}. However, its difficult for traditional GANs to fully understand the target domain knowledge with limited number of samples, which often lead to the poor translation quality. 
DDPM resolves this question by using a large-scale pre-training with text-to-image data~\cite{saharia2022palette,ruiz2023dreambooth,ramesh2022hierarchical} and integrating multimodal information like large-scale language models~\cite{shao2023prompting,lan2023improving}. 

Given a target prompt, the diffusion model has the ability to generate realistic yet eligible content. This attractive property has considerable potential for realizing the image translation task with fewer or zero target samples. 

\begin{figure}[t]
\centering
\includegraphics[width=0.9\linewidth]{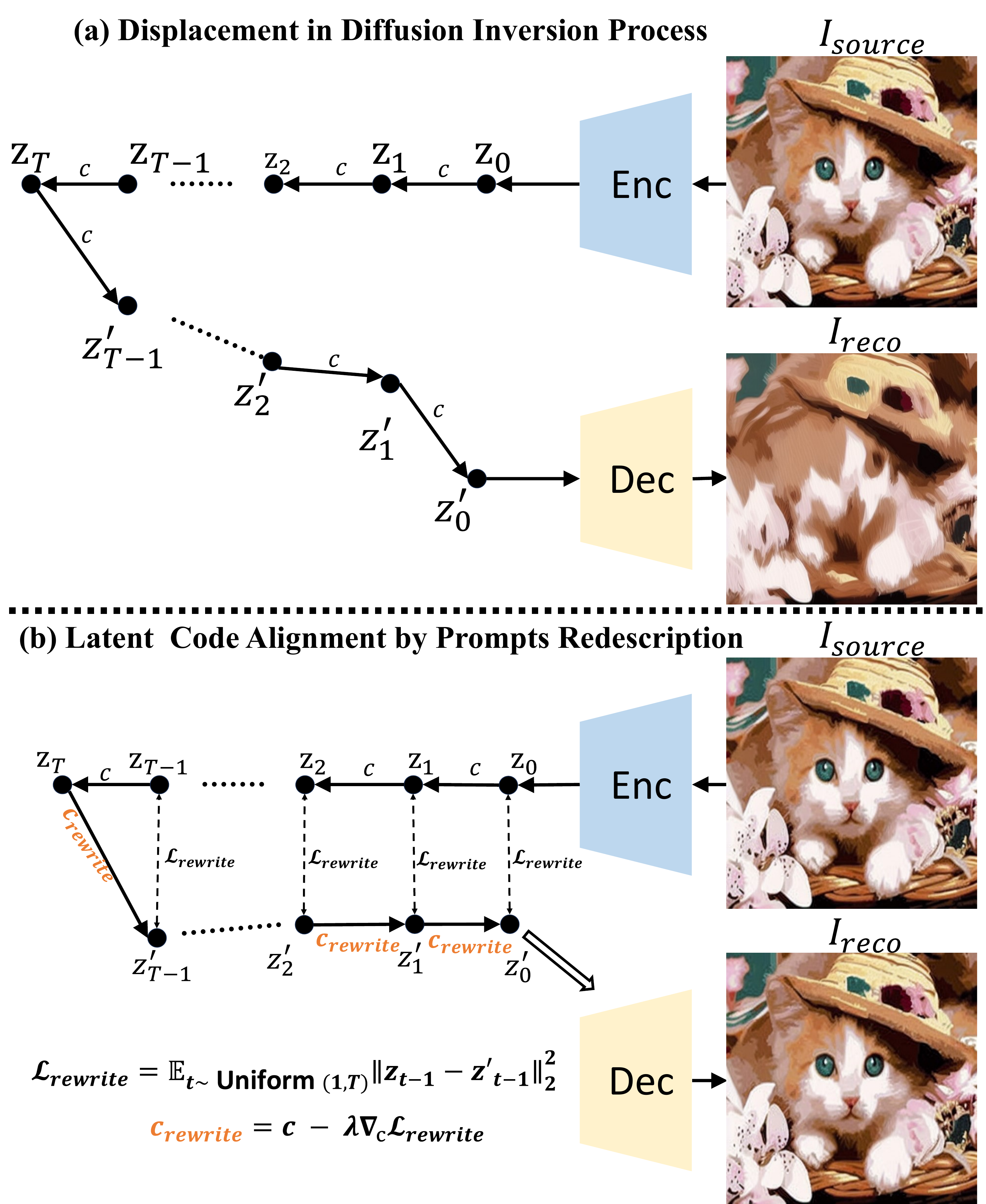}
\caption{{To show displacement effect, the reconstruction process of typical DDIM work~\cite{parmar2023zero} is visualized in Fig.\ref{fig:rec_res} (a), which can be formulated as: $z_0\to z_T\to z'_0$. However, errors accumulate in typical diffusion methods, causing biases in latent codes $[z_0, z'_0]$ and deviations in $[I_{source}, I_{reco}]$. To align the latent codes, we propose a prompt redescription mechanism to realize a mirror effect between the source and reconstructed image in the diffusion model (MirrorDiffusion). }}
\label{fig:rec_res}
\vspace{-4mm}
\end{figure}

\begin{figure*}[t]
\centering
\includegraphics[width=0.9\linewidth]{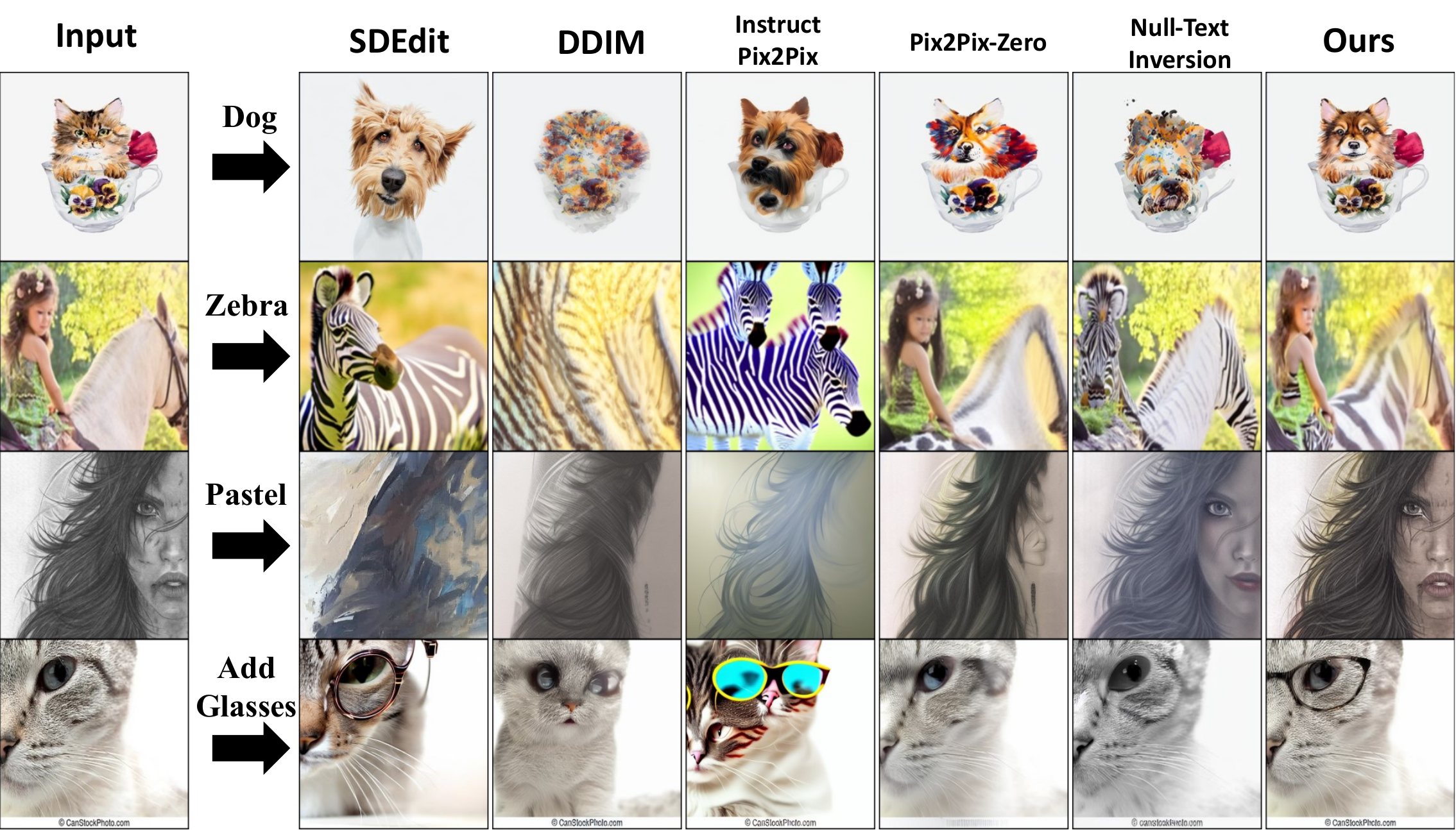}
\caption{{Visualization results. Compared with state-of-the-art diffusion approaches across four tasks, our method excels in generating highly realistic translation results with \textbf{excellent structure consistency.}}}

\label{fig:exp_res}
\vspace{-4mm}
\end{figure*}

To address zero-shot image translation, Pix2Pix-Zero~\cite{parmar2023zero} first adopts the diffusion pipeline. Specifically, Pix2Pix-Zero measures the distance between source and target domain sentences by applying a CLIP~\cite{radford2021learning} model. By utilizing that domain gap between text embedding, Pix2Pix-Zero successfully achieves zero-shot image-to-image translation with a pre-trained Denoising Diffusion Implicit Models (DDIM)~\cite{song2020denoising}. Literally, typical DDIM usually converts the image into latent code, and then performs a latent code re-sampling w.r.t target domain prompt to demonstrate zero-shot image translation. This pipeline is able to generate images of the target domain without target images. As shown in Fig.~\ref{fig:rec_res}, the generated results often deviate from the structure of the source domain, which violates the structure consistency. To keep the structure of the generated results consistent with the original content, many efforts~\cite{hertz2022prompt,mokady2022null} are devoted to investigating the reconstruction pipeline in DDIM. Prompt2Prompt~\cite{hertz2022prompt} first reconstructs the image structure at the early sampling stage with the original prompt, and then applies a word swap mechanism to change the prompt for image detail generation. Instruct Pix2Pix~\cite{brooks2022instructpix2pix} proposes a triple dataset, which contains caption, edit instruction and edited instruction for image editing. To keep the structural consistency, Instruct Pix2Pix utilizes a cross-attention mechanism to ensure the edited image becomes consistent with the original image. Null-text Inversion~\cite{mokady2022null} optimizes the null-text embedding for classifier-free guidance to ensure that the reconstruction results consistent with the source image. This null-text optimization requires the high precision of provided text prompts, otherwise, the details of the generated results will be skewed. SDEdit~\cite{meng2021sdedit} realizes fidelity and consistency in the image translation by using stochastic differential equation (SDE) for image encoding and decoding. Since the SDE is invertible, the inversion process from noise to image is naturally realized. 

However, the forementioned methods gradually accumulate displacement during the diffusion and inversion processes, which makes the reconstructed results gradually deviate from the source image. As shown in Fig.~\ref{fig:rec_res}, the reconstructed image appears a displacement effect, which further affects the accuracy of the translation result. To solve the displacement effect in image reconstruction, we propose a prompt redescription strategy to realize a mirror effect between the source and reconstructed image during the diffusion process (MirrorDiffusion). Specifically, we address the deviation problem of the reconstructed image by aligning the text prompts and latent codes at each time step of the reconstruction process. With the revised DDIM inversion, MirrorDiffusion obtains accurate target text embedding and latent code for zero-shot image translation. Our contributions can be summarized as:

\begin{itemize}
\item A prompt redescription mechanism is proposed to address the displacement problem of image reconstruction in DDIM Inversion. With the prompt redescription, we achieve a reliable yet effective image reconstruction.
\item Based on the revised DDIM inversion, we align the latent code with the text prompt during the diffusion process to further ensure consistency in zero-shot image translation.
\item Extensive experiments demonstrate that MirrorDiffusion achieves superior performance over the state-of-the-art diffusion models on zero-shot image translation benchmarks by clear margins and practical model stability.
\end{itemize}

\section{Methodology}
\subsection{Displacement in Diffusion Inversion Process}
In Denoising Diffusion Implicit Models (DDIM) and its derivatives~\cite{song2020denoising,parmar2023zero}, the sampling and inversion processes~\cite{song2020denoising,dhariwal2021diffusion} are stochastic, and thus the inversion process often fail to reconstruct the input as shown in Fig.~\ref{fig:rec_res} (a). Specifically, given a source domain image $I_{source}$, we first apply the encoder $Dec(\cdot)$ of the diffusion model~\cite{song2020denoising} to convert it into a latent code $z_0$. And then send $z_0$ into DDIM for diffusion, the object function is as follow:

\begin{equation}
    L_{DDIM}=\min _{\theta} E_{t \sim \text { U }(1, T)}\left \|N_{gaus}-\epsilon_{\theta}\left(z_{t}, t, c  \right) \right \|_{2}^{2},
\label{eq:ddim}
\end{equation}
$N_{gaus} \sim \mathcal{N}(0,1)$ is Gaussian noise with standard normal distribution, $t$ is time step in DDIM with a range of $\left [ 1,T \right ] $, $c$ represents the embedding of a conditional text prompt, $\epsilon_\theta$ is the noise prediction network in DDIM, $z_t$ represents the latent code in high dimensional space after $t$ times of diffusion process. The physical meaning of Equ.~\ref{eq:ddim} is to minimize the difference between the noise predicted by $\epsilon_{\theta}$ and the real distribution standard Gaussian noise. 

After DDIM optimization, as show in Fig.~\ref{fig:rec_res} (a), we can input the latent code $z_T$ into the DDIM for stepwise sampling and image reconstruction:

\begin{equation}
    z'_{T-1} = Sample\left(\epsilon_\theta,z_T,c_T,T\right),
\label{eq:sample}
\end{equation}

where the $Sample\left(\cdot \right)$ represents the stepwise sampling of DDIM, $\epsilon_\theta$ is the noise prediction network in DDIM. 

As shown in Fig.~\ref{fig:rec_res} (a), with $T$ steps sampling, we can convert $z_T$ into $z'_0$, and obtain the reconstructed image by: $I_{reco}= Dec\left(z'_{0}\right) $. However, in the inversion process, $\epsilon_\theta$ may deviate from the standard Gaussian noise, and this displacement will gradually accumulate on $\left [ Z_T, Z'_{T-1},..,Z'_0 \right ]$, resulting in an collapse in $I_{reco}$.

To show this displacement effect and collapse in $I_{reco}$, we visualize reconstruction process of some typical DDIM works~\cite{parmar2023zero,song2020denoising}. As shown in Fig.~\ref{fig:rec_res} (a), the diffusion inversion process realizes the reconstruction by: $z_0\to z_T\to z'_0$. However, the distribution between $[z_0, z'_0]$ appears a displacement effect, resulting in a deviation between $[I_{source}, I_{reco}]$. 

\subsection{Prompts Redescription}
To make the reconstruction explicit, we propose a prompts redescription strategy. Specifically, in the reconstruction phrase as shown in Fig.~\ref{fig:rec_res} (b), we calculate the difference between the latent codes during inversion and reconstruction as:
\begin{equation}
\mathcal{L}_{rewrite} =\min _{\theta} E_{ t \sim \text { Uniform }(1, T)}\left\| z_{t-1} - z'_{t-1} \right\|^2_2,
\label{eq:RePrompts}
\end{equation}

\begin{figure*}[t]
\centering
\includegraphics[width=0.9\linewidth]{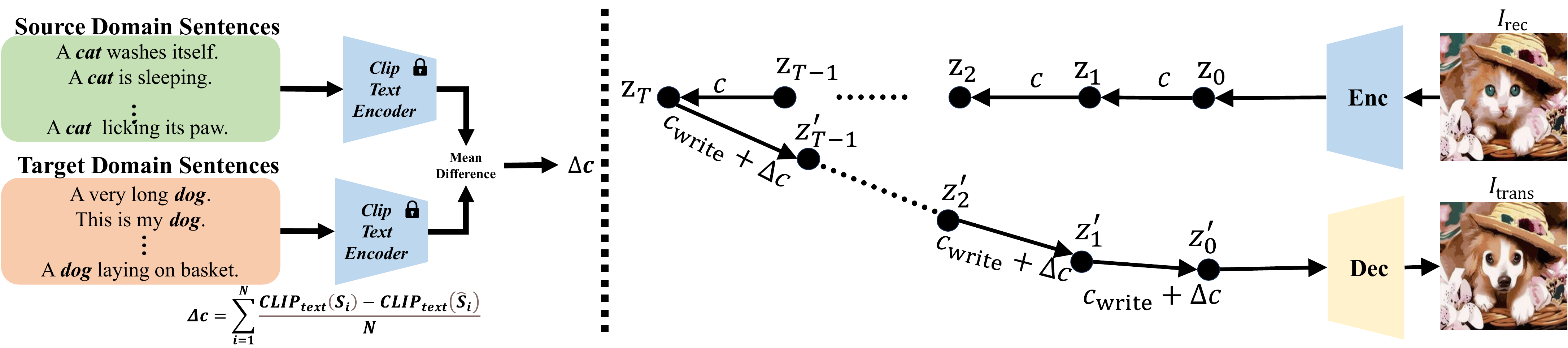}
\caption{{The framework overview of MirrorDiffusion. With the prompt redescription mechanism, our model obtains the firmly aligned $[z_0, z'_0]$, $[I_{source}, I_{reco}]$ combinations. We apply CLIP~\cite{radford2021learning} to compute the domain gap $\Delta c$ between the source domain and target domain for image editing. Specifically, the CLIP~\cite{radford2021learning} is used to extract the high-level features of source domain sentences and target domain sentences, respectively. And the mean difference, which is computed along those features, is represented as the domain gap $\Delta c$. Then, we apply the target text embedding $c_{rewrite} + \Delta c$ for zero-shot image translation with diffusion inversion process. With $T$-time inversion, MirrorDiffusion can obtain the corresponding latent code $z'_0$, which corresponds to $I_{trans}$ with $Dec(\cdot)$. }}
\label{fig:edit_pipe}
\vspace{-4mm}
\end{figure*}
where the $z'_{t-1}$ indicates the sampling noise of the current time step, which is obtained by: $z'_{t-1} = Sample\left(\epsilon_\theta,z_t,c_{rewrite},t\right)$. Then, we use the $\mathcal{L}_{rewrite}$ to implement a redescription toward current text prompt $c_t$ as:
\begin{equation}
c_{rewrite} = c_t - \lambda \nabla_c \mathcal{L}_{rewrite},
\label{eq:Redescription}
\end{equation}
where $\nabla_c$ takes the partial derivative on $c$ and obtain gradient for prompt redescription. According the rewritten text prompt $c_{rewrite}$, we re-sample $z_t$ at each time step in inversion process as: $z'_{t-1} = Sample\left(\epsilon_\theta,z_t,c_{rewrite},t\right)$. As shown in Fig.~\ref{fig:rec_res} (b), to our surprise, the prompt redescription mechanism ensures that $[z_0, z'_0]$, $[I_{source}, I_{reco}]$ are firmly aligned after $T$ steps sampling.

\subsection{Zero-shot Image Translation with MirrorDiffusion}
With the prompts redescription mechanism, we can obtain the aligned combination $[c_{rewrite}, z'_0]$ toward $\epsilon_\theta$. As shown in Fig.~\ref{fig:exp_res}, our model can implement zero-shot image translation by further editing $c_{rewrite}$ according to the target domain. 

As shown in Fig.~\ref{fig:edit_pipe}, we apply CLIP~\cite{radford2021learning} to compute the domain gap $\Delta c$ between the source domain and target domain. Specifically, the CLIP~\cite{radford2021learning} is used to extract the high-level features of source domain sentences and target domain sentences, respectively. And the mean difference, which is computed along those features, is represented as the domain gap $\Delta c$. 

As shown in Fig.~\ref{fig:edit_pipe}, we then use the updated target text embedding $c_{rewrite} + \Delta c$ for a latent code sampling as: 
\begin{equation}
z'_{t-1} = Sample\left(\epsilon_\theta,z'_t,c_{rewrite}+\Delta c,t\right),
\label{eq:sampling2}
\end{equation}
where $\Delta c$ represents the target domain direction, such as $Dog \to Cat$. And $c_{rewrite}$ is the source domain text embedding, which was firmly aligned by the prompt redescription mechanism. After $T$-time sampling, our model can obtain the corresponding latent code $z'_0$. As shown in Fig.~\ref{fig:edit_pipe}, we can easily obtain translated image with a renewed $z'_0$ as:

\begin{equation}
I_{trans} = Dec(z'_0),
\label{eq:decoder}
\end{equation}
where $I_{trans}$ represents the translated image, $Dec(\cdot)$ is an image decoder~\cite{kingma2013auto}. 

\section{Experiment}

\subsection{Datasets and Metric}

To evaluate the gap between our model and baselines, we selected three sub-datasets from the LAION-5B dataset~\cite{schuhmann2022laion} containing cat, horse and sketch images, each containing 250 images. Subsequently, we formulated the following image transformation tasks, including, Translate Cat to Dog(C2D-F), Add Glasses to Cat(C2G-F), Translate Sketch to Oil Pastel(S2O-F), Translate Horse to Zebra(H2Z-F). 

We conducted a comparative analysis of performance differences between our approach and several baselines, including: SDEdit~\cite{meng2021sdedit}, DDIM~\cite{song2020denoising}, InstructPix2pix~\cite{brooks2022instructpix2pix}, Pix2Pix-Zero~\cite{parmar2023zero} and NULL-Text inversion~\cite{mokady2022null} methods, across these datasets.For quantitative quality evaluation, we used CLIP-ACC~\cite{hessel2021clipscore} ,Structure Dist~\cite{tumanyan2022splicing} , Structure Similarity Index Measure~\cite{wang2004image} (SSIM) and Learned Perceptual Image Patch Similarity~\cite{zhang2018unreasonable} (LPIPS), which are evaluated in terms of whether the translation is successful.
\begin{figure*}[ht]
    \centering
    \includegraphics[width=0.92\linewidth]{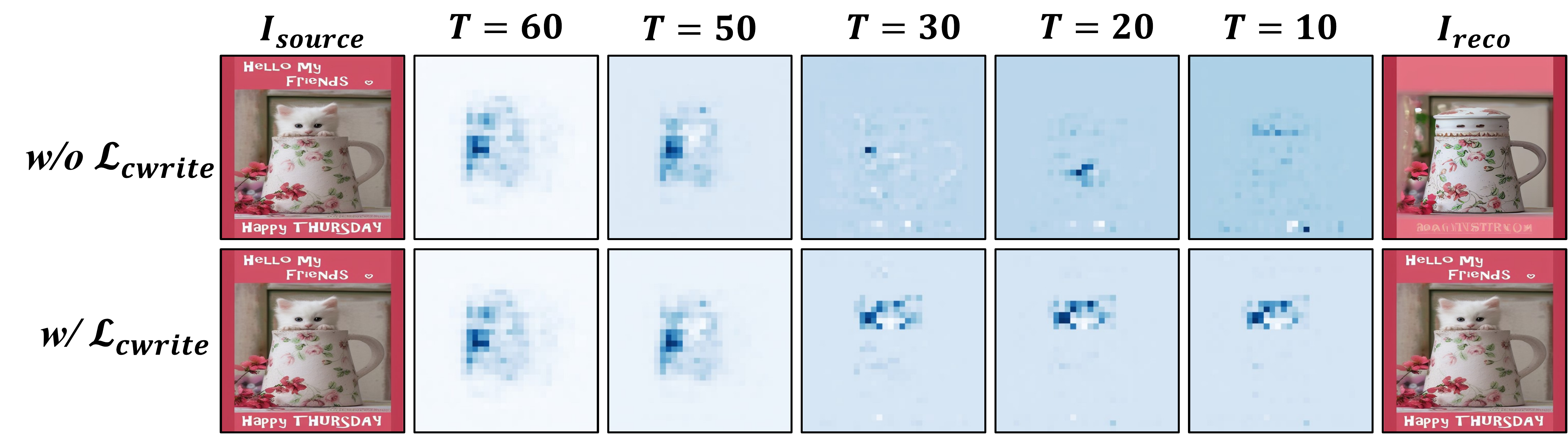}
    \caption[something short]{Attention maps of `w/o $\mathcal{L}_{rewrite}$' and `w/ $\mathcal{L}_{rewrite}$' during reconstruction process.}
    \label{fig:attention_map}
\end{figure*}

\begin{table*}[t]
\caption{Comparison of quantitative results. We evaluate the results in terms of the four evaluation metrics CLIP-ACC, Structure Dist, SSIM, and LPIPS, with the pink cells indicating the results of the best performance and the orange cells indicating the results of the second best performance}
\vspace{-2mm}
\centering
    \tiny
        \begin{tabular}{ccccccccccccccccc}
    \hline
                                   & \multicolumn{4}{c}{C2D-F}                                  & \multicolumn{4}{c}{H2Z-F}                                  & \multicolumn{4}{c}{C2G-F}                                           & \multicolumn{4}{c}{S2O-F}                                      \\ \cmidrule(lr){2-5} \cmidrule(lr){6-9}  \cmidrule(lr){10-13} \cmidrule(lr){14-17} 
    \multirow{-2}{*}{METHOD}        & Clip$\uparrow$                        & Structure$\downarrow$        & SSIM$\uparrow$                                    & LPIPS$\downarrow$          & Clip$\uparrow$                         & Structure$\downarrow$        & SSIM$\uparrow$                                    & LPIPS$\downarrow$            & Clip$\uparrow$                        & Structure$\downarrow$            & SSIM$\uparrow$                                    & LPIPS$\downarrow$        & Clip$\uparrow$                         & Structure$\downarrow$                & SSIM$\uparrow$                                    & LPIPS$\downarrow$    \\ \hline
    \multicolumn{1}{c}{SDEdit~\cite{meng2021sdedit}}     & 66.9                                   & 0.146                                 & 0.441                                 & 0.552                                 & 78.7 & 0.223                                 & 0.441                                  & 0.573                                 & 76.9                                 & 0.133                                 & 0.428                                 & 0.566                                 & 56.1                                 & 0.133                                 & 0.502                                 & 0.519                                                                \\ 
    
    \multicolumn{1}{c}{DDIM~\cite{song2020denoising}}        & 60.2                                   & 0.127                                 & 0.637                                 & 0.439                                 & 72.5                                 & 0.159                                 & 0.6441                                 & 0.491                                 & 68.8                                 & 0.114                                 & 0.589                                 & 0.435                                 & 53.5                                 & 0.122                                 & 0.633                                 & 0.441                                    \\ 
    \multicolumn{1}{c}{Instruct Pix2Pix~\cite{brooks2022instructpix2pix}}       & 72.5                                   & 0.086                                 & 0.699                                 & 0.275                                 & 76.4                                 & 0.256                                 & 0.455                                  & 0.711                                 & 74.4                                 & 0.155                                 & 0.342                                 & 0.633                                 & 66/3                                 & 0.130                                 & 0.649                                 & 0.485                                                                   \\ 
    \multicolumn{1}{c}{Pix2Pix-Zero~\cite{parmar2023zero}}    & {\cellcolor{Khaki}\textbf{75.2}} & {\cellcolor{Khaki}\textbf{0.071}} & {0.718} & {0.272} & 78.3                                 & {\cellcolor{Khaki}\textbf{0.106}} & {\cellcolor{Khaki}\textbf{0.671}}  & {0.385} & {80.3} & {\cellcolor{Khaki}\textbf{0.047}} & 0.653 & 0.246 & {\cellcolor{LightPink}\textbf{70.7}} & {\cellcolor{Khaki}\textbf{0.059}} & 0.741 & {\cellcolor{Khaki}\textbf{0.240}} \\ 
     \multicolumn{1}{c}{NULL-Text-Inversion~\cite{mokady2022null} } & 74.5 & 0.081 & \cellcolor{Khaki}\textbf{0.72} & \cellcolor{Khaki}\textbf{0.25} & \cellcolor{Khaki}\textbf{81.6}
 & 0.1369 & 0.625 & 0.36  & \cellcolor{Khaki}\textbf{83.2}  & 0.054 & \cellcolor{Khaki}\textbf{0.743} & \cellcolor{Khaki}\textbf{0.225} & 69.7 & 0.066 & \cellcolor{Khaki}\textbf{0.743} & 0.248 \\
    \hline
    
    \multicolumn{1}{c}{Ours } & \cellcolor{LightPink}{\textbf{77.0}} & {\cellcolor{LightPink} \textbf{0.04}} & {\cellcolor{LightPink} \textbf{0.782}} & {\cellcolor{LightPink} \textbf{0.15}} & {\cellcolor{LightPink} \textbf{82.1}} & {\cellcolor{LightPink} \textbf{0.0758}} & {\cellcolor{LightPink} \textbf{0.722}} & {\cellcolor{LightPink} \textbf{0.271}} & {\cellcolor{LightPink} \textbf{83.9}} & {\cellcolor{LightPink} \textbf{0.024}} & {\cellcolor{LightPink} \textbf{0.797}} & {\cellcolor{LightPink} \textbf{0.13}} & {\cellcolor{LightPink} \textbf{70.7}} & {\cellcolor{LightPink} \textbf{0.043}} & {\cellcolor{LightPink} \textbf{0.768}} & {\cellcolor{LightPink} \textbf{0.176}} \\ \hline
    \end{tabular}
\label{tab:exp_res}
\vspace{-3mm}
\end{table*}
\subsection{ Implementation details} 
During the image editing process, we optimize only the variables in the text rewrite process and the pre-trained stable diffusion model remains in a frozen state. In this experiment, we established the weight parameter $\lambda$ for $\mathcal{L}_{rewtire}$ as 1, employing the Adam optimizer to update the value of $c_{rewrite}$, with a learning rate of 0.0001. In this paper, the number of iterations for the DDIM inversion, as well as for the DDIM editing sampling and reconstruction sampling are both set to 60. In the image editing process, we use classifier-free guidance~\cite{ho2022classifier} to predict the noise at each time step. All inputs and generated results are with a size of 512$\times$512$\times$3.

\begin{figure}[t]
\centering
\includegraphics[width=0.9\linewidth]{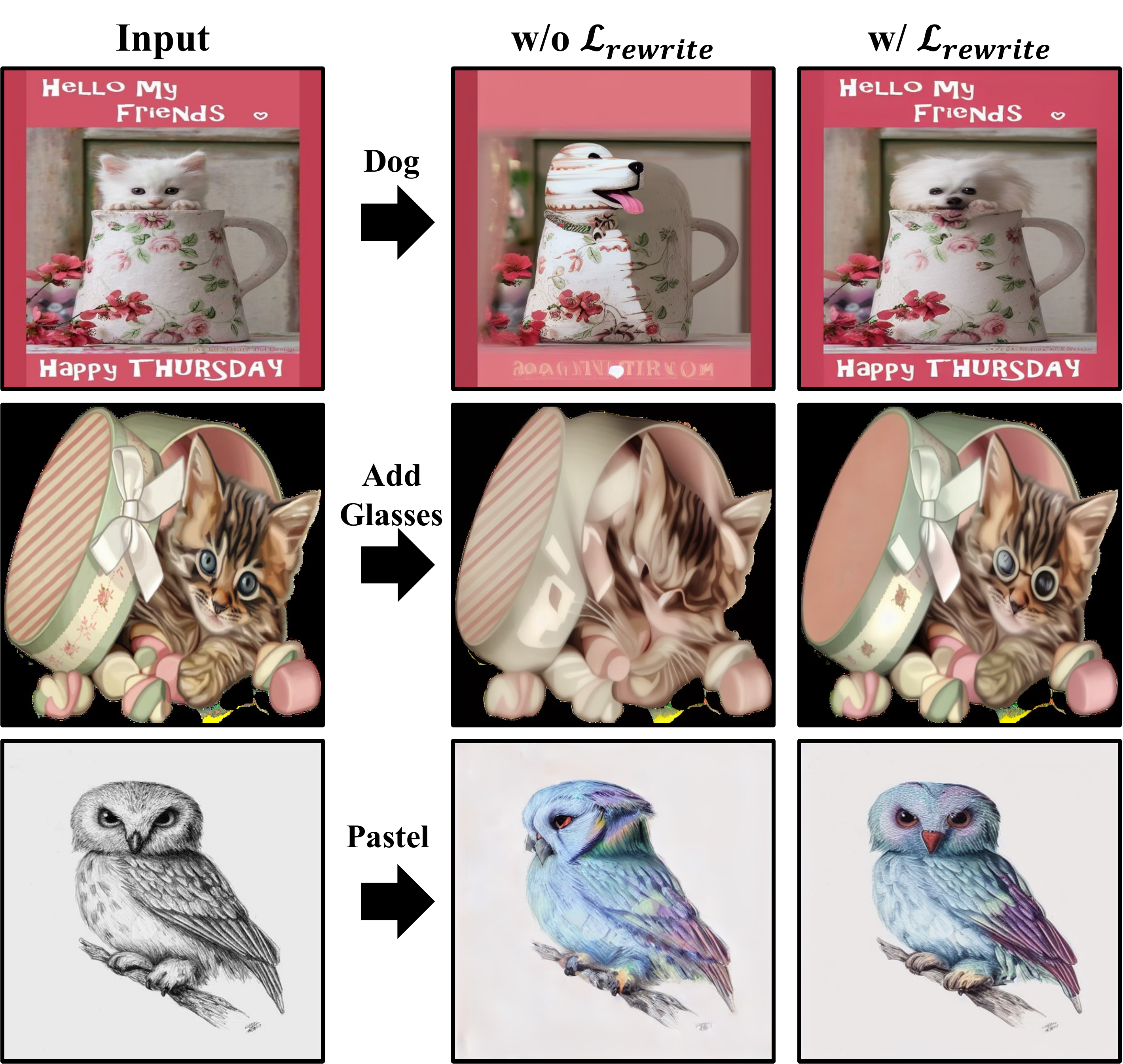}
\caption{Ablation study of $\mathcal{L}_{rewrite}$. It can be observed that $\mathcal{L}_{rewrite}$ plays a significant role in preserving the appearance.}
\label{fig:ab_res}
\vspace{-4mm}
\end{figure}

\begin{table}[t]
\centering
\caption{Ablation on $\mathcal{}L_{rewrite}$. 
The results show that the structure can be well preserved with $\mathcal{}L_{rewrite}$.
}
\vspace{-2mm}
\label{tab:ab_exp}
\begin{tabular}{ccccc}
\hline
\multirow{2}{*}{} & \multicolumn{4}{c}{C2G-F}    \\ \cline{2-5}
                  & Clip$\uparrow$ & structure$\downarrow$ & SSIM $\uparrow$ & LPIPS $\downarrow$ \\ \hline
w/o $\mathcal{L}_{rewrite}$               & 82.6  & 0.055      & 0.729 & 0.255 \\ \hline
w/ $\mathcal{L}_{rewrite}$               & \cellcolor{LightPink} \textbf{83.9}  & \cellcolor{LightPink} \textbf{0.024}      & \cellcolor{LightPink} \textbf{0.797} & \cellcolor{LightPink} \textbf{0.132} \\ \hline
\end{tabular}
\vspace{-4mm}
\end{table}

\subsection{Comparison}
As shown in Fig.~\ref{fig:exp_res}, we compare the visual appearance of our approach with the baseline models on the four tasks. It can be seen that SDEdit, DDIM and Instruct Pix2Pix struggle to maintain appearance consistency, Pix2Pix-Zero and Null-Text Inversion perform well in terms of general appearance, but fall short in terms of preserving details and translation accuracy. In contrast, our method performs well in both appearance preservation and translation correctness. Our quantitative evaluation results also achieved the best results. In Tab.~\ref{tab:exp_res}, the pink cells represent the best results and the orange cells represent the second-best results. It can be clearly seen that we achieved the best results in both translation quality and structure preservation. 
To demonstrate the effectiveness of our approach on diverse tasks, we supplement three additional tasks with different scenarios: w/o glasses $\rightarrow$ w/ glasses, Male $\rightarrow$ Female and Fox $\rightarrow$ Dog. The image data for these tasks are sourced from CelebA-HQ~\cite{karras2017progressive} and AFHQ~\cite{choi2020stargan}. As shown in Fig.~\ref{fig:add_exp}, our method effectively achieves high-quality image translation in these tasks.

\subsection{Ablation Study}

\textbf{Effects of $\mathcal{L}_{rewrite}$ during reconstruction.} As shown in Fig.~\ref{fig:attention_map}, we show the attention maps during the reconstruction process of MirrorDiffusion. To verify the effectiveness of our method, we show the attention maps of `w/ $\mathcal{L}_{rewrite}$' and `w/o $\mathcal{L}_{rewrite}$'. As shown in Fig.~\ref{fig:attention_map}, without $\mathcal{L}_{rewrite}$, the attention map of the cat gradually deviates from itself, leading to a poor reconstruction result. 
With the proposed $\mathcal{L}_{rewrite}$, the cat's attention maps center on critical regions and complete a faithful reconstruction.

\textbf{Effects of $\mathcal{L}_{rewrite}$ during editing.} To show the effect of $\mathcal{L}_{rewrite}$ during editing, we present a comparison between the results obtained using `w/ $\mathcal{L}_{rewrite}$' and `w/o $\mathcal{L}_{rewrite}$'. As depicted in Fig.~\ref{fig:ab_res}, the differences between these two approaches are clearly visible. Without the use of $\mathcal{L}_{rewrite}$, the edited results are difficult to effectively preserve the original appearance. Conversely, when $\mathcal{L}_{rewrite}$ is employed, the results show improved preservation of the original appearance while accomplishing the desired image translation.

\textbf{Quantitative results analysis.} Furthermore, we also present some quantitative metrics in the C2GF task. As shown in Tab.~\ref{tab:ab_exp}, it can be observed that our image translation results demonstrate significant improvements across these four metrics, which highlights the effectiveness of the prompt redescription mechanism.

\begin{figure}[h]
    \centering
    \includegraphics[width=0.9\linewidth]{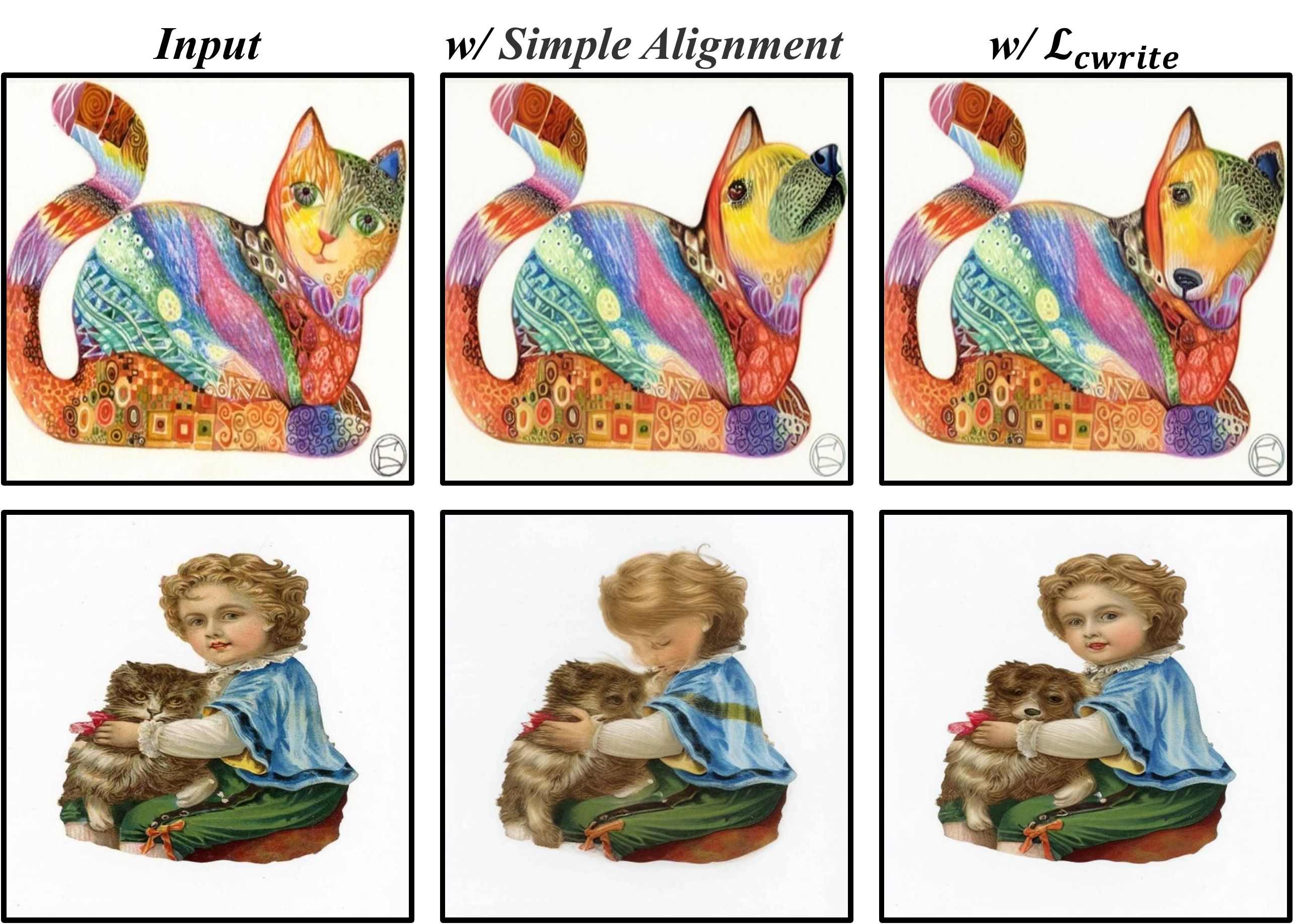}
    \caption[something short]{Comparison of the generation results of `w/ Simple Alignment' and `w/ $\mathcal{L}_{rewrite}$'.}
    \label{fig:simple_align}
\end{figure}

\textbf{Effects of prompt rewrite module.} To verify the effectiveness of the alignment and rewrite modules, we have made an ablation study. Suppose we apply an independent alignment module instead of using the rewrite module, we show the generated results in Fig.~\ref{fig:simple_align}. During the editing stage, there exists a gap between the prompt and expected prompt at each time step. A simple alignment strategy fails to compensate this gap among prompts, leading to a poor quality. Although an independent alignment module aligns the overall structure with the original image, it fails to preserve image details. As shown in Fig.~\ref{fig:simple_align}, without rewrite module, the orientation of boy's face and dog's head exhibit an unreasonable rotation.
Instead, our method aligns the prompt with expected prompt at each time step by rewrite module, achieving a successfully image translation that preserves the structural consistency.

\begin{figure}[t]
    \centering
    \includegraphics[width=0.7\linewidth]{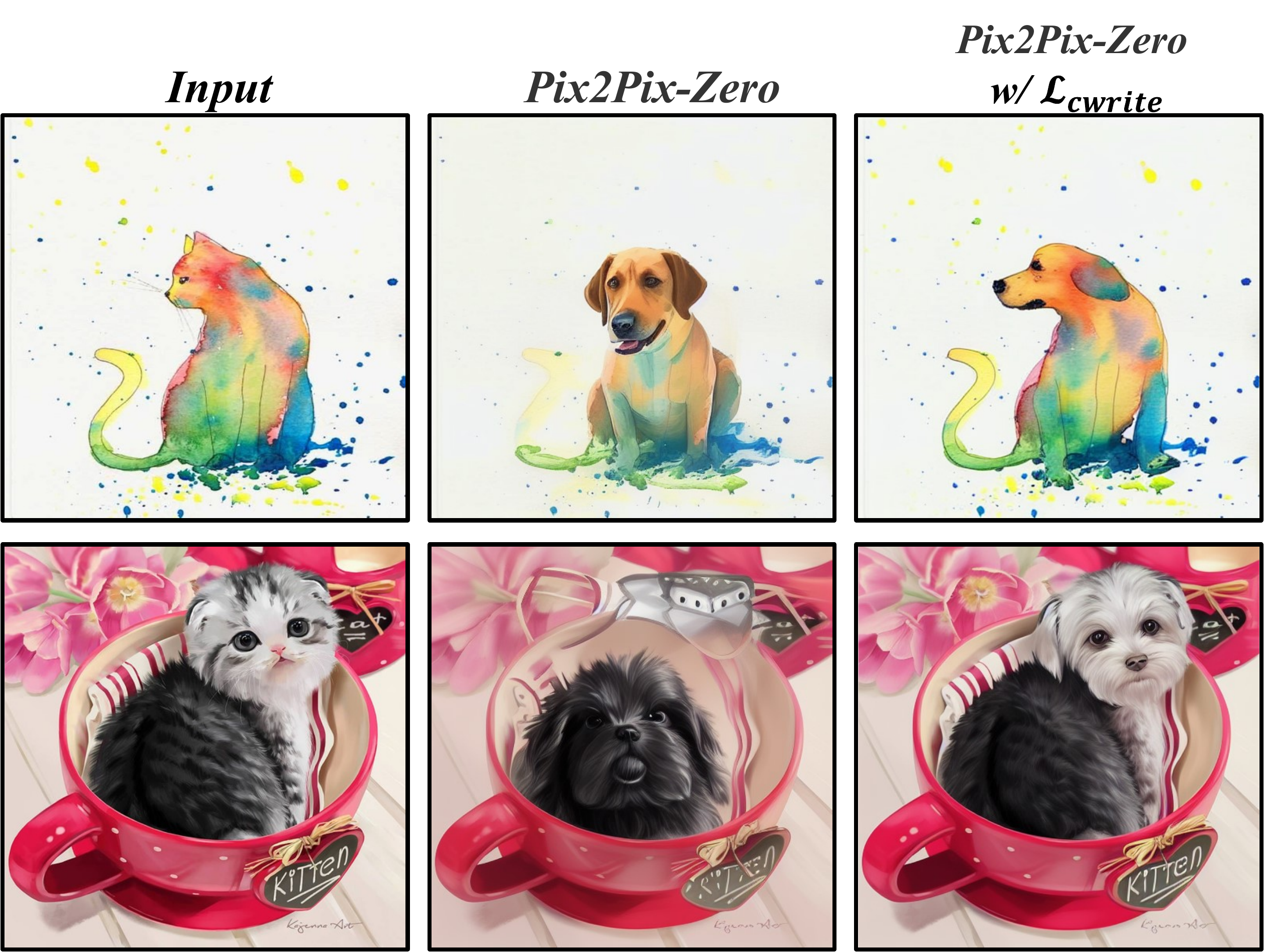}
    \caption[something short]{The generated results of Pix2Pix-Zero `w/o $\mathcal{L}_{rewrite}$' and `w/ $\mathcal{L}_{rewrite}$'.}
    \label{fig:p2p}
\end{figure}

\begin{figure}[h]
    \centering
    \includegraphics[width=0.7\linewidth]{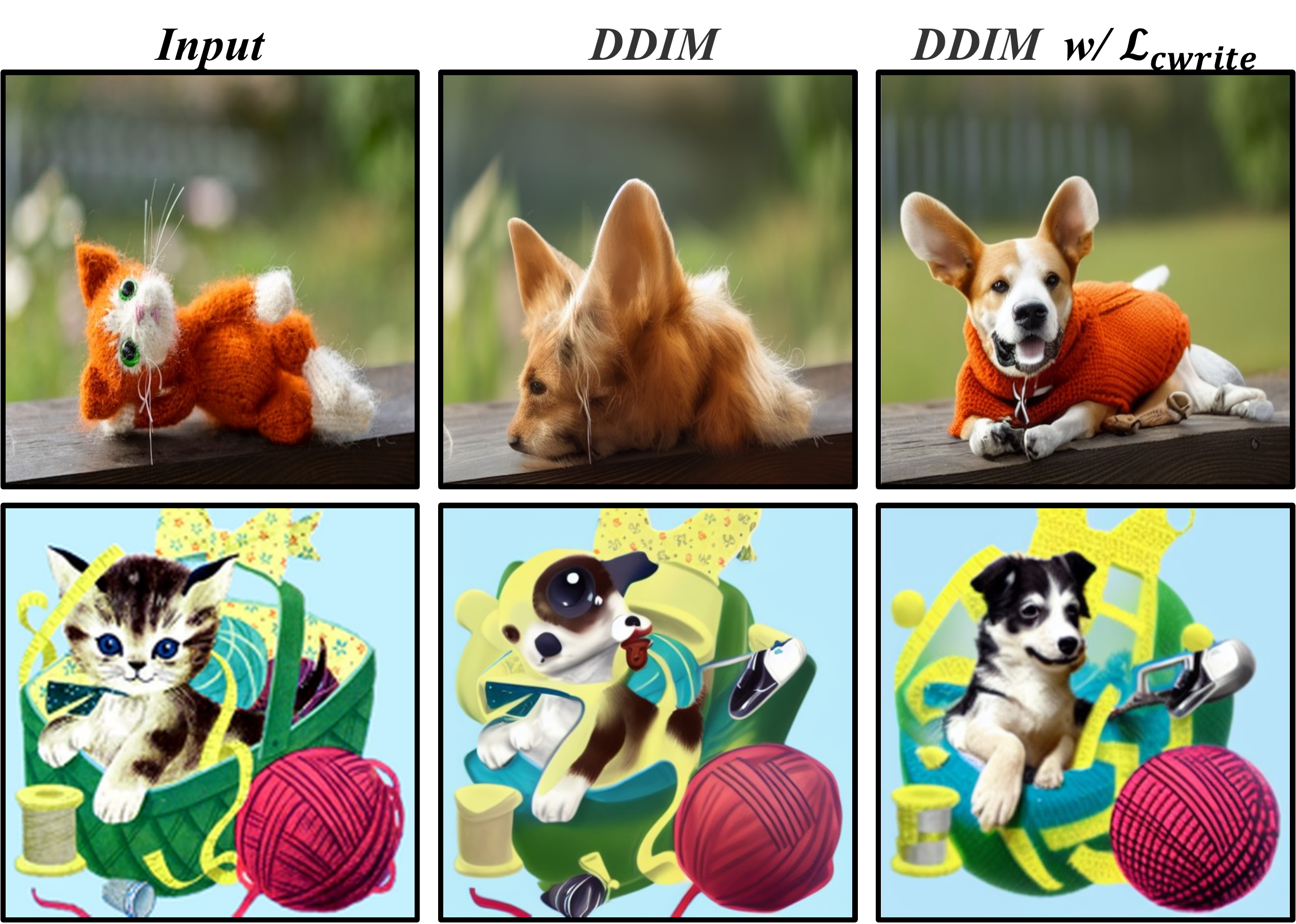}
    \caption[something short]{ The generated results of DDIM `w/o $\mathcal{L}_{rewrite}$' and `w/ $\mathcal{L}_{rewrite}$'.}
    \label{fig:ddim}
\end{figure}

\textbf{Further investigations on the rewrite module.} We apply the prompt rewrite module to  DDIM~\cite{song2020denoising} and Pix2Pix-Zero~\cite{parmar2023zero} for further investigation. As shown in Fig.~\ref{fig:p2p}, the structure of `Pix2Pix-Zero w/ $\mathcal{L}_{rewrite}$' is consistent to the original image. Additionally, we show the results of DDIM with the proposed rewrite module. As shown in Fig.~\ref{fig:ddim}, `DDIM w/ $\mathcal{L}_{rewrite}$' is more faithful to the structure of the original image.

\section{Conclusion}

In this paper, we aims to address the deviation and displacement problems of current text-to-image diffusion models in image translation tasks. By introducing a new prompt redescription mechanism, our method surpasses state-of-the-art diffusion-based image translation methods on both visual results and quantitative results.

\bibliographystyle{IEEEtran}
\bibliography{egbib}

\end{document}